%% file: main.tex
\documentclass{article}
\usepackage[final]{neurips_2023}
\usepackage[utf8]{inputenc} 
\usepackage[T1]{fontenc}    
\usepackage{hyperref}       
\usepackage{url}            
\usepackage{booktabs}       
\usepackage{amsfonts}       
\usepackage{amsmath}
\usepackage{nicefrac}       
\usepackage{microtype}      
\usepackage{xcolor}         
\usepackage[pdftex]{graphicx}
\usepackage{multirow}
\usepackage{enumitem}
\usepackage{algorithm}
\usepackage{algorithmic}
\usepackage{caption}

\setcitestyle{numbers,open={[},close={]}}

\title{Microscaling Data Formats for \\ Deep Learning}

\vspace{-1.5em}
 \author{\hspace{-1em}
 \footnotesize{Bita Darvish Rouhani\thanks{email correspondence: birouhan@microsoft.com} \> Ritchie Zhao \> Ankit More \> Mathew Hall \> Alireza Khodamoradi \> Summer Deng} 
 \\
 \footnotesize{\textbf{Dhruv Choudhary} \> \textbf{Marius Cornea} \> \textbf{Eric Dellinger} \> \textbf{Kristof Denolf} \> \textbf{Stosic Dusan} \> \textbf{Venmugil Elango}} \\  
 \footnotesize{\textbf{Maximilian Golub} \> \textbf{Alexander Heinecke} \> \textbf{Phil James-Roxby} \> \textbf{Dharmesh Jani} \>  \textbf{Gaurav Kolhe}} 
 \\  
 \footnotesize{\textbf{Martin Langhammer} \> \textbf{Ada Li} \> \textbf{Levi Melnick} \> \textbf{Maral Mesmakhosroshahi} \> \textbf{Andres Rodriguez}}  
 \\ 
 \footnotesize{\textbf{Michael Schulte} \> \textbf{Rasoul Shafipour}  \> \textbf{Lei Shao} \> \textbf{Michael Siu} \> \textbf{Pradeep Dubey} \> \textbf{Paulius Micikevicius}}
 \\
 \footnotesize{\textbf{Maxim Naumov} \> \textbf{Colin Verrilli} \> \textbf{Ralph Wittig} \> \textbf{Doug Burger} \>\textbf{Eric Chung}} \\\\
 \footnotesize{\textit{Microsoft}\quad \textit{AMD}\quad \textit{Intel}\quad \textit{Meta}\quad \textit{NVIDIA}\quad \textit{Qualcomm Technologies Inc.}}
 }

\begin{document}

\maketitle
\vspace{-1em}
\begin{abstract}
  Narrow bit-width data formats are key to reducing the computational and storage costs of modern deep learning applications. This paper evaluates Microscaling (MX) data formats that combine a per-block scaling factor with narrow floating-point and integer types for individual elements.
  MX formats balance the competing needs of hardware efficiency, model accuracy, and user friction.
  Empirical results on over two dozen benchmarks demonstrate practicality of MX data formats as a drop-in replacement for baseline FP32 for AI inference and training with low user friction. We also show the first instance of training generative language models at sub-8-bit weights, activations, \textit{and} gradients with minimal accuracy loss and no modifications to the training recipe.
\end{abstract}

\section{Introduction} \label{sec:intro}
Recent advances in AI capabilities such as conversational question answering,
intelligent code completion, and text-to-image generation have seen rapid adoption in practical technologies.
These advances have been realized primarily through scaling up the size of the underlying deep learning model.
However, this scaling up has led to a significant increase in the computing power and storage capacity necessary to train and deploy such models.

One method to reduce deep learning models' computational and storage cost is to use low bit-width data formats instead of the conventional FP32.
Great strides have been made to enable training using FP16, Bfloat16, and most recently FP8~\cite{ocp_fp8}, as well as to perform inference in narrow integer formats like INT8.
Native support for low bit-width formats is now commonplace in AI-oriented hardware such as GPUs, TPUs, and edge inference devices.
The narrowest formats, such as FP8 and INT8, require per-tensor scaling factors to adjust to the dynamic range of each tensor. Tensor level scaling has has been shown to be insufficient, though, for sub-8-bit formats due to their limited dynamic range.
Research has shown that micro scaled data formats that associate scaling factors with fine-grained sub-blocks of a tensor are more effective in sub-8 bit regime (e.g., \cite{drumond2018hbfp, darvish2020pushing, dai2021vs, darvish2023mx}).

This paper evaluates Microscaling (MX) data formats~\cite{mxspec} --- the first open standard for a family of micro-scaled datatypes aimed at deep learning training and inference. The MX standard aims to create an effective data format by achieving a balance among three key factors:
\begin{itemize}
\vspace{-0.5em}
    \item \textbf{Hardware Efficiency} --- Maximize compute and storage efficiency via reduced bit-width.
    \item  \textbf{Model Accuracy} --- Minimize the gap in the quality of results compared with baseline FP32 for AI training and inference.
    \item  \textbf{User Friction} --- Ensure seamless integration within existing training and inference frameworks and generalizability across different workloads.
\end{itemize}
Details on the MX standard and the concrete binary formats can be found in the OCP Microscaling Specification~\cite{mxspec}. This paper will focus on the empirical results of using MX formats for direct-cast inference, error diffusion inference, and finetuned inference, as well as training on various benchmarks.
Our results corroborate the effectiveness of MX formats in balancing the competing demands of hardware efficiency, model accuracy, and user friction.
8-bit MX formats can perform inference directly on FP32 pretrained models with minimal accuracy loss and without the need for calibration or finetuning.
Inference with 6-bit MX formats is also very close to FP32 after quantization-aware fine-tuning or using a post-training quantization method.
Using 6-bit MX formats, we demonstrate the first instance of training large transformer models with sub-8-bit weights, activations, and gradients to an accuracy matching FP32 without modifications to the training recipe.
Going even further, we show that training of large transformers can be done with 4-bit MX format weights, incurring only a minor accuracy drop.

The custom CUDA library to emulate MX formats on existing GPUs can be found at~\cite{mxlib_link}. This library can be used to reproduce the experimental results reported in this paper.

\section{Microscaling} \label{sec:mx}
\begin{figure}[hp]
  \centerline{\includegraphics[width=0.7\linewidth]{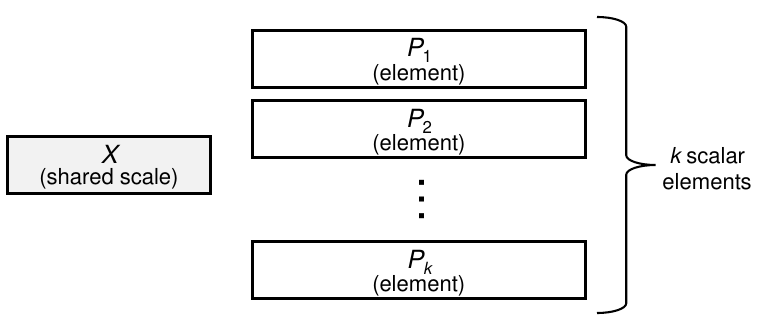}}
  \caption{A single block in a Microscaling data format.
           The block encodes a vector of $k$ numbers, each with value $XP_i$.}
  \label{fig:mx_format}
\end{figure}

A basic unit of data in an MX format represents a vector of $k$ numbers and consists of a single \textit{shared scale} $X$ and $k$ scalar \textit{elements} $\{P_i\}_{i=1}^k$ (see Figure~\ref{fig:mx_format}). This unit of data is called an MX block and is defined by the combination of \textit{block size} $k$, scale data format, and element data format. The two data formats are independent of one another, and all $k$ elements share the same element data format.
The layout of an MX block is not prescribed --- an implementation may store $X$ contiguously with or separately from the elements.

Let $\{v_i\}_{i=1}^k$ be the $k$ real numbers represented in an MX block. The value of each number can be inferred as follows:
\begin{itemize}[nosep]
\setlength{\itemsep}{0pt}
    \item If $X = \text{NaN}$, then $v_i = \text{NaN}$ for all $i$
    \item If $|XP_i| > Vmax_{Float32}$ then $v_i$ is implementation-defined
    \item Otherwise, $v_i = XP_i$ 
\end{itemize}
where $Vmax_{Float32}$ refers to the largest representable magnitude in IEEE Float32. 

\subsection{Special Value Encodings}
MX formats can encode NaN in up to two ways. First: if $X$ is NaN, then all $k$ values in the MX block is NaN regardless of the encodings of $P_i$. Second: if $X$ is not NaN, each element $P_i$ may individually encode NaN.

Depending on the element format, MX formats can encode Inf by letting $X$ be a number (i.e., not a NaN) and each $P_i$ individually encode Inf. The shared scale $X$ does not encode Inf.

\subsection{Concrete MX Formats}
Table~\ref{tab:concrete_formats} shows the parameters that define the concrete MX formats, which are named by prepending "MX" to the name of the element data format. All concrete MX formats use E8M0 (an 8-bit exponent) as the format for the shared scale. The representable exponents of these formats is a superset of the representable exponents of FP32.

Details on the FP8 element data formats can be found in the OCP FP8 specification~\cite{ocp_fp8}. Details on the other element data formats and the E8M0 scale format can be found in the OCP Microscaling Specification~\cite{mxspec}.

\input{tab_concrete}

\vspace{-1em}
\section{Scalar Float to MX Format Conversion} \label{sec:conversion}
In this paper, we use Algorithm~\ref{alg:float2mx} for conversion from scalar floating-point format (e.g., FP32) to an MX format. This algorithm follows the semantics outlined in Section 6.3 of the OCP Microscaling Specification~\cite{mxspec}, and is provided as a working example. Note that, the specification allows for other implementation-defined conversion recipes --- i.e., conversion to MX formats is \textit{not necessarily required} to follow Algorithm $1$. 

\begin{algorithm}
\caption{Convert vector of scalar floats $\{V_i\}_{i=1}^k$ to an MX block
$\{X,\>\{P_i\}_{i=1}^k\}$}
\label{alg:float2mx}
\begin{algorithmic}[1]
\REQUIRE $emax_{elem} = \text{exponent of the largest normal number in the element data format}$
\STATE 
$shared\_exp \gets \lfloor \log_2(\max_{i}(|V_i|)) \rfloor - emax_{elem}$
\STATE $X \gets 2^{shared\_exp}$
\FOR{$i=1$ to $k$}
  \STATE $P_i = quantize\_to\_element\_format(V_i / X)$, clamping normal numbers
\ENDFOR
\RETURN $X,\>\{P_i\}_{i=1}^k$
\end{algorithmic}
\end{algorithm}

On Line 1, $shared\_exp$ contains an offset of $emax\_elem$ to map the max input exponent to the largest binade in the element data format. This enables full utilization of the element data format's exponent range. 

On Line 4, when quantizing $V_i/X$, normal numbers that exceed the representable range of the element format are clamped to the maximum representable value, preserving the sign. Infs and NaNs are not clamped. This is in accordance with the OCP MX specification.

On Line 4, $P_i$ is set to zero if the corresponding input $V_i$ is a subnormal Float32 number. This is not described in the OCP MX specification and was done to simplify the algorithm.

When converting multi-dimensional tensors, a principle axis must be selected for the shared scale (typically the reduction dimension in matrix multiplication). For a 2D matrix, the scale can be shared by every $k$ element in a row or column. Transposing a 2D matrix in an MX format changes the axis of the shared scale --- i.e., conversion to MX format and transposing are not commutative operations. 

\section{Experimental Results} \label{sec:experiment}

\subsection{Compute Flow} \label{sec:flow}
\begin{figure}[htb]
  \centerline{\includegraphics[width=1.0\linewidth]{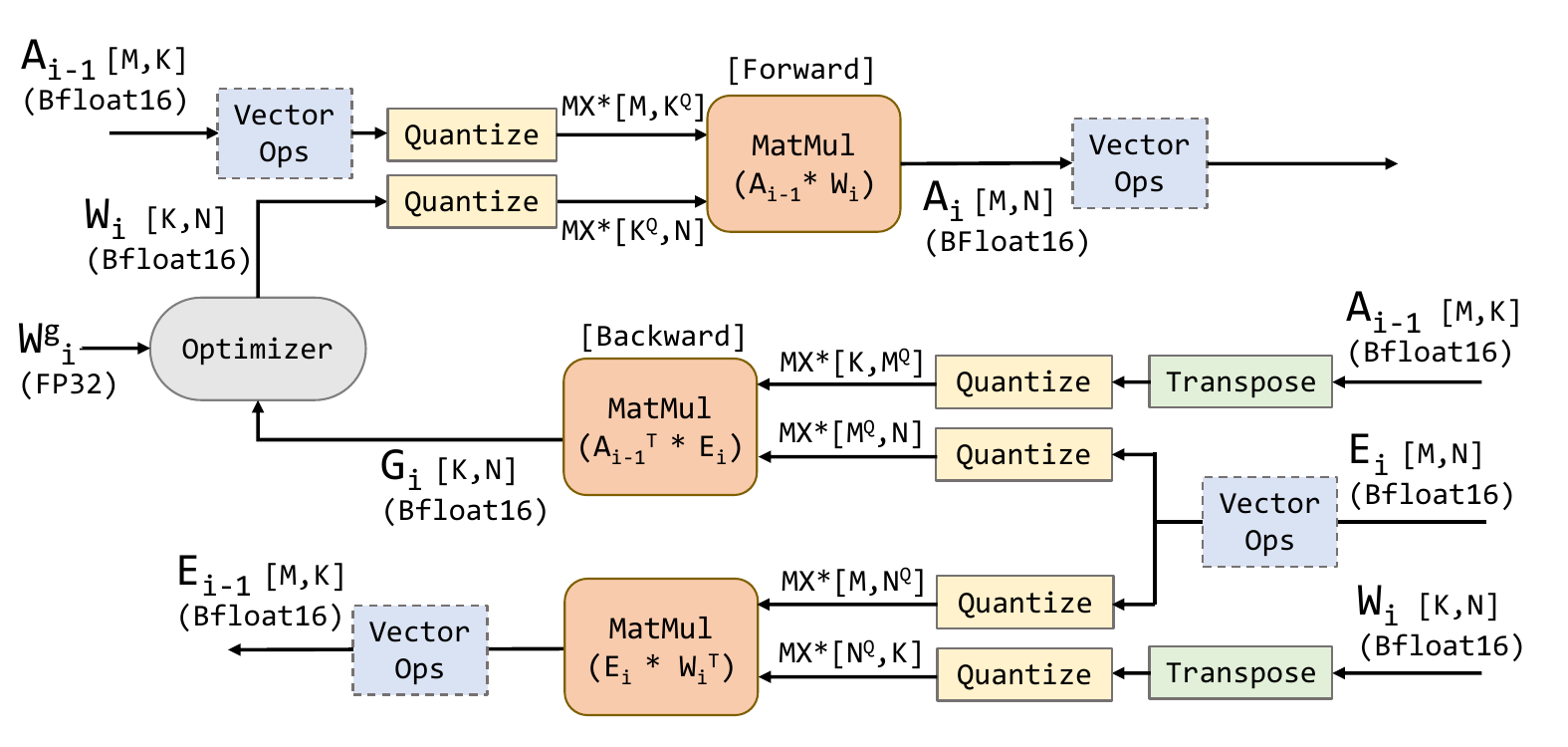}}
  \caption{Compute flow with MX formats (denoted as MX*). In the diagram, MatMul includes any dot product operation such as matmul, linear, and convolution. Vector Ops include non-dot product operations like activations, normalization, Softmax, and residual add.}
  \label{fig:mx_flow}
\end{figure}

Figure~\ref{fig:mx_flow} shows an example compute flow for training using an MX format. 
For operations involving dot products (e.g., matmul and convolution) in both forward and backward passes, the two inputs are converted to MX format, and the operation is performed using the efficient dot product from Section 6.2 of the OCP Microscaling Specification~\cite{mxspec}.
Vector operations (e.g., layernorm, Softmax, GELU, and residual add) are performed in a scalar floating-point format like Bfloat16 or FP32.
The dot product operations produce outputs in the scalar float format.
A master copy of the weights is kept in FP32, and this copy is updated in each training step. In all the training examples in this paper, we use the compute flow illustrated in Figure~\ref{fig:mx_flow}.

Due to non-commutative nature of transpose and quantization into MX formats (see Section~\ref{sec:conversion}), the quantized weights $W_i$ and their transpose $W_i^T$ must be stored as two separate tensors.
Note that the two tensors do not need to be stored in working memory simultaneously unless a very fine-grained interleaving of the forward and backward passes is employed.

\subsection{Methodology}
We used a custom library to emulate MX formats on existing GPUs. The library is implemented as a custom CUDA extension in PyTorch and performs quantization following Figure~\ref{fig:mx_flow}. In particular, we explored four settings:

\begin{itemize}
    \item \textit{Direct-cast Inference}. The quantized inference is performed on a trained FP32 model. All GeMMs in the forward pass are quantized unless explicitly called out otherwise (the backward pass is not executed at all).
    \item \textit{Error Diffusion Inference}. The error diffusion algorithm is a Post Training Quantization (PTQ) algorithm derived from GPFQ ~\cite{GPFQ_UCSD}. It performs quantization using a small calibration dataset. In this experiment, all activations and weights in the forward pass are quantized to the same format for simplicity. This PTQ process is a quick one-pass process without a training loop or needing any tuning parameter. 
    \item \textit{Finetuned Inference}. Quantization-aware finetuning is done on a trained FP32 model for a small number of epochs. For this fine-tuning, all GeMMs in the forward pass are quantized, while the backward pass is performed in FP32. Hyperparameter exploration is used to find proper finetuning hyperparameters. 
    \item \textit{Training}. A model is trained from scratch using a compute flow where all GeMMs in both forward and backward passes are quantized (see Figure~\ref{fig:mx_flow}. For mixed-precision training where the weights and activations use different data formats, the gradients ($\text{E}_{\text{i}}$ in Figure~\ref{fig:mx_flow}) are quantized to the activation format.
\end{itemize}

Our benchmark suite contains two types of tasks: discriminative and generative. 

\subsection{Discriminative Inference}
In this section, we examine inference results with MX formats across a variety of discriminative tasks including language translation, text encoding, image classification, speech recognition, and recommendation models. Table~\ref{tab:disc-inference} summarizes the results related to \textbf{direct-cast inference}. Results for \textbf{finetuned inference} are reported in Table~\ref{tab:disc-finetune}, and results for \textbf{PTQ with error diffusion inference} are reported in Table~\ref{tab:disc-ed}.

In these experiments, the same MX formats were used for both weights and activations following Algorithm~\ref{alg:float2mx}. Round-half-to-nearest-even was used for conversion to MX formats. The results presented in Table~\ref{tab:disc-inference} corroborates the effectiveness of MXINT8 as a drop-in replacement for FP32 with minimal accuracy drop.  
For MXFP8 and MXFP6, the general trend is that the variant of the format with more mantissa bits was better for direct-cast inference.
With finetuned inference (Table~\ref{tab:disc-finetune}), MXFP6\_E2M3 is able to achieve close-to-parity with FP32.

\input{Experiment_results/discriminative_inference}
\vspace{-1em}
\input{Experiment_results/discriminative_error_diffusion}
\vspace{-1em}
\input{Experiment_results/discriminative_finetune}

\subsection{Generative Inference}
We leveraged the open source LM Eval Harness by Eleuther AI for our evaluation of MX data formats in generative inference of OpenAI GPT3-175B and open source LLaMA-7B.\footnote{https://github.com/EleutherAI/lm-evaluation-harness/tree/1736d78dd9615107e68ec7f74043b02d4ab68d12.} All benchmarks were run under zero-shot settings (i.e., no examples were presented to the models before evaluation). Our benchmark suite includes the following subset:

\textbf{Lambada} --- Lambada is a long range prediction task, where the model must predict the last word in a long narrative passage. We used the version of lambada data used to evaluate GPT2 in LM Harness.

\textbf{Wikitext} --- The wikitext task is based on the wikitext-2 dataset and requires the model to predict long sequences based on high quality Wikipedia articles.
GPT3-175B was not evaluated on this task as Wikipedia data was part of its training corpus~\cite{browngpt3}.

\textbf{ARC dataset} --- The Arc tasks are both multiple choice tasks consisting of nearly $8000$ science exam questions, with the dataset split into easy and more challenging questions. The model is tasked with picking the correct answer from several options. 

\textbf{Hendryck's Test} --- Hendryck's test suite is a set of tasks that measure how knowledgeable a model is in $57$ different fields. We used \textbf{computer science}, \textbf{international law}, and \textbf{jurisprudence} as a subset for this study. These tasks are all multiple choice questions, where the model must pick the correct answer from the options presented.

Table~\ref{tab:gpt3-inference} and Table~\ref{tab:llama-inference} show results for direct-cast inference on OpenAI GPT3-175B~\cite{browngpt3} and open source LLaMA-7B, respectively. 
Due to the size of these models, no quantization-aware finetuning was performed. The columns with a single MX format use that format for both weights and activations; the other columns list separate formats for weights (Wt) and activations (Act) and utilize mixed-precision.

MXINT8 matched baseline FP32 to within the standard deviation on all tasks for both GPT3-175B and LLaMA-7B. MXINT8 once again proves to be a compelling alternative to FP32 for low-friction direct-cast inference.

\input{Experiment_results/generative_inference}

\subsection{Generative Training}

Table~\ref{tab:GPT-mxfp6-training} and Figure ~\ref{fig:gpt_mxfp6_training} show the language model loss obtained from training GPT-like models of various size (20M-1.5B) using MXFP6\_e3m2 for both the forward and backward passes (see Figure~\ref{fig:mx_flow}).
The training is done using the ADAM optimizer, with hyperparameters tuned for FP32. The same hyperparameters were reused for the MX format runs with no changes. All the models are trained to efficiency with number of steps calculated based on the scaling power-laws~\cite{kaplan2020scaling}. Round-half-away-from-zero rounding was used for conversion to MX formats.

The results in Table~\ref{tab:GPT-mxfp6-training} and Figure ~\ref{fig:gpt_mxfp6_training} show that MXFP6\_e3m2 is capable of delivering a model quality matching that of FP32 at much lower circuitry footprint.
\textbf{MXFP6 provides the first demonstration of training generative language models to parity with FP32 using 6-bit weights, activations, and gradients with no modification to the training recipe.}

Pushing the limits even further, Table~\ref{tab:GPT-mixed_training} and Figure~\ref{fig:gpt_mixed_training} show the results from training the same GPT-like models, this time under a mixed-precision setting with MXFP4 weights and MXFP6\_e3m2 activations. The gradients used the same data format as the activations.
The training hyperparameters were the same as before.
\textbf{Our results demonstrate that generative language models can be trained with MXFP4 weights and MXFP6 activations and gradients incurring only a minor penalty in the model loss.} This is once again with no modifications to the training recipe.

\input{Experiment_results/gen_training}
\input{Experiment_results/gen_training_mixed}

\section{Conclusion} \label{sec:conclusion}
This paper evaluates MX data formats that integrate a block-level scale on top of narrow bit-width elements.
The evaluated concrete MX formats provide compelling alternatives to FP32 training and inference with minimal user friction.
Experimental results show the effectiveness of MX formats for a variety of deep learning models including generative language models, image classification, speech recognition, recommendation models, and translation.

In particular, MXINT8 is a compelling drop-in replacement to FP32 for low-friction direct-cast inference.
MXFP6 closely matches FP32 for inference after quantization-aware finetuning.
MXFP6 also, for the first time, enables generative language model training at sub-8-bit weights, activations, and gradients without sacrificing model accuracy or needing changes to the training recipe.
Reducing the bit-width even further, we showcase training with MXFP4 weights and MXFP6 activations and gradients, incurring only a minor loss penalty for generative language models.

 \section*{Acknowledgment}
 The authors would like to thank the following individuals for their invaluable support and contributions: Ian Bratt, Nigel Stephens, Jelena Milanovic, John Brothers, Yuan Yu, Rani Borkar, Saurabh Dighe, Brian Harry, Matt Perry, Renee L'Heureux, Dimitry Melts, Jasmine Klar, and Steve Scott. 

\small
\bibliographystyle{unsrtnat}
\bibliography{main}

\end{document}

%% file: tab_concrete.tex
\begin{table}[h]
\caption{Concrete MX-compliant data formats and their parameters.}
\begin{center}

\begin{tabular}{|c|c|c|c|c|c|}
\hline
\textbf{Format} & \textbf{Block} & \textbf{Scale} & \textbf{Scale} & \textbf{Element} & \textbf{Element} \\
\textbf{Name} & \textbf{Size} & \textbf{Data Format} & \textbf{Bits} & \textbf{Data Format} & \textbf{Bit-width} \\
\hline
\hline
MXFP8  & 32 & E8M0 & 8 & FP8 (E4M3 / E5M2) & 8 \\
\hline
MXFP6  & 32 & E8M0 & 8 & FP6 (E2M3 / E3M2) & 6 \\
\hline
MXFP4  & 32 & E8M0 & 8 & FP4 (E2M1) & 4 \\
\hline
MXINT8 & 32 & E8M0 & 8 & INT8 & 8 \\
\hline
\end{tabular}
\label{tab:concrete_formats}
\end{center}
\end{table}

%% file: Experiment_results/discriminative_inference.tex
\begin{table}[h]
\centering
\caption{Direct-cast inference with MX data formats. For each experiment, the FP32 baseline was quantized (both weights and activations) with no additional tweaks. MXINT8 is a compelling alternative to FP32 for low-friction direct-cast inference.\\}
\label{tab:disc-inference}
\resizebox{\textwidth}{!}{%
\renewcommand{\arraystretch}{1.4}
\begin{tabular}{ccccc|ccccccc|}
\hline
\multicolumn{1}{|c|}{\multirow{2}{*}{Task}} &
  \multicolumn{1}{c|}{\multirow{2}{*}{Family}} &
  \multicolumn{1}{c|}{\multirow{2}{*}{Model}} &
  \multicolumn{1}{c|}{\multirow{2}{*}{Dataset}} &
  \multirow{2}{*}{Metric} &
  \multicolumn{1}{c|}{\multirow{2}{*}{\begin{tabular}[c]{@{}c@{}}Baseline\\ FP32\end{tabular}}} &
  \multicolumn{1}{c|}{\multirow{2}{*}{MXINT8}} &
  \multicolumn{2}{c|}{MXFP8} &
  \multicolumn{2}{c|}{MXFP6} &
  \multirow{2}{*}{MXFP4} \\ \cline{8-11}
\multicolumn{1}{|c|}{} &
  \multicolumn{1}{c|}{} &
  \multicolumn{1}{c|}{} &
  \multicolumn{1}{c|}{} &
   &
  \multicolumn{1}{c|}{} &
  \multicolumn{1}{c|}{} &
  \multicolumn{1}{c|}{E4M3} &
  \multicolumn{1}{c|}{E5M2} &
  \multicolumn{1}{c|}{E2M3} &
  \multicolumn{1}{c|}{E3M2} &
   \\ \hline
\multicolumn{1}{|c|}{\multirow{3}{*}{\begin{tabular}[c]{@{}c@{}}Language \\ Translation\end{tabular}}} &
  \multicolumn{1}{c|}{\multirow{2}{*}{\begin{tabular}[c]{@{}c@{}}Transformers\\ (Enc-Dec)\end{tabular}}} &
  \multicolumn{1}{c|}{Transformer-Base~\cite{transformer}} &
  \multicolumn{1}{c|}{\multirow{2}{*}{WMT-17}} &
  \multirow{3}{*}{\begin{tabular}[c]{@{}c@{}}BLEU\\ Score ↑\end{tabular}} &
  \multicolumn{1}{c|}{26.85} &
  \multicolumn{1}{c|}{26.64} &
  \multicolumn{1}{c|}{26.27} &
  \multicolumn{1}{c|}{25.75} &
  \multicolumn{1}{c|}{26.38} &
  \multicolumn{1}{c|}{25.97} &
   22.68 \\ \cline{3-3} \cline{6-12} 
\multicolumn{1}{|c|}{} &
  \multicolumn{1}{c|}{} &
  \multicolumn{1}{c|}{Transformer-Large~\cite{transformer}} &
  \multicolumn{1}{c|}{} &
   &
  \multicolumn{1}{c|}{27.63} &
  \multicolumn{1}{c|}{27.56} &
  \multicolumn{1}{c|}{27.44} &
  \multicolumn{1}{c|}{27.02} &
  \multicolumn{1}{c|}{27.52} &
  \multicolumn{1}{c|}{27.22} &
  26.33 \\ \cline{2-4} \cline{6-12} 
\multicolumn{1}{|c|}{} &
  \multicolumn{1}{c|}{LSTM} &
  \multicolumn{1}{c|}{GNMT~\cite{gnmt}} &
  \multicolumn{1}{c|}{WMT-16} &
   &
  \multicolumn{1}{c|}{24.44} &
  \multicolumn{1}{c|}{24.52} &
  \multicolumn{1}{c|}{24.53} &
  \multicolumn{1}{c|}{24.45} &
  \multicolumn{1}{c|}{24.51} &
  \multicolumn{1}{c|}{24.44} &
  23.75 \\ \hline
\multicolumn{1}{|c|}{\multirow{2}{*}{\begin{tabular}[c]{@{}c@{}}Language \\ Encoding\end{tabular}}} &
  \multicolumn{1}{c|}{\multirow{2}{*}{\begin{tabular}[c]{@{}c@{}}Transformers \\ (Enc-Only)\end{tabular}}} &
  \multicolumn{1}{c|}{BERT-Base~\cite{gptbert}} &
  \multicolumn{1}{c|}{\multirow{2}{*}{Wikipedia}} &
  \multirow{2}{*}{\begin{tabular}[c]{@{}c@{}}F-1\\ Score ↑\end{tabular}} &
  \multicolumn{1}{c|}{88.63} &
  \multicolumn{1}{c|}{88.58} &
  \multicolumn{1}{c|}{88.47} &
  \multicolumn{1}{c|}{87.04} &
  \multicolumn{1}{c|}{88.38} &
  \multicolumn{1}{c|}{88.05} &
 84.94 \\ \cline{3-3} \cline{6-12} 
\multicolumn{1}{|c|}{} &
  \multicolumn{1}{c|}{} &
  \multicolumn{1}{c|}{BERT-Large~\cite{gptbert}} &
  \multicolumn{1}{c|}{} &
   &
  \multicolumn{1}{c|}{93.47} &
  \multicolumn{1}{c|}{93.41} &
  \multicolumn{1}{c|}{93.42} &
  \multicolumn{1}{c|}{93.32} &
  \multicolumn{1}{c|}{93.45} &
  \multicolumn{1}{c|}{93.27} &
  90.97 \\ \hline
\multicolumn{1}{|c|}{\multirow{5}{*}{\begin{tabular}[c]{@{}c@{}}Image \\ Classification\end{tabular}}} &
  \multicolumn{1}{c|}{\multirow{2}{*}{\begin{tabular}[c]{@{}c@{}}Vision \\ Transformer\end{tabular}}} &
  \multicolumn{1}{c|}{DeiT-Tiny~\cite{deit}} &
  \multicolumn{1}{c|}{\multirow{5}{*}{\begin{tabular}[c]{@{}c@{}}ImageNet\\ ILSVRC12\end{tabular}}} &
  \multirow{5}{*}{\begin{tabular}[c]{@{}c@{}}Top-1\\ Acc. ↑\end{tabular}} &
  \multicolumn{1}{c|}{72.16} &
  \multicolumn{1}{c|}{72.20} &
  \multicolumn{1}{c|}{71.37} &
  \multicolumn{1}{c|}{70.11} &
  \multicolumn{1}{c|}{71.56} &
  \multicolumn{1}{c|}{70.16} &
  56.72 \\ \cline{3-3} \cline{6-12} 
\multicolumn{1}{|c|}{} &
  \multicolumn{1}{c|}{} &
  \multicolumn{1}{c|}{DeiT-Small~\cite{deit}} &
  \multicolumn{1}{c|}{} &
   &
  \multicolumn{1}{c|}{80.54} &
  \multicolumn{1}{c|}{80.56} &
  \multicolumn{1}{c|}{79.83} &
  \multicolumn{1}{c|}{79.00} &
  \multicolumn{1}{c|}{80.11} &
  \multicolumn{1}{c|}{79.04} &
  71.35 \\ \cline{2-3} \cline{6-12} 
\multicolumn{1}{|c|}{} &
  \multicolumn{1}{c|}{\multirow{3}{*}{CNN}} &
  \multicolumn{1}{c|}{ResNet-18~\cite{resnet}} &
  \multicolumn{1}{c|}{} &
   &
  \multicolumn{1}{c|}{70.79} &
  \multicolumn{1}{c|}{70.80} &
  \multicolumn{1}{c|}{69.08} &
  \multicolumn{1}{c|}{66.16} &
  \multicolumn{1}{c|}{69.71} &
  \multicolumn{1}{c|}{66.10} &
  48.77 \\ \cline{3-3} \cline{6-12} 
\multicolumn{1}{|c|}{} &
  \multicolumn{1}{c|}{} &
  \multicolumn{1}{c|}{ResNet-50~\cite{resnet}} &
  \multicolumn{1}{c|}{} &
   &
  \multicolumn{1}{c|}{77.40} &
  \multicolumn{1}{c|}{77.27} &
  \multicolumn{1}{c|}{75.94} &
  \multicolumn{1}{c|}{73.78} &
  \multicolumn{1}{c|}{76.42} &
  \multicolumn{1}{c|}{73.75} &
  42.39 \\ \cline{3-3} \cline{6-12} 
\multicolumn{1}{|c|}{} &
  \multicolumn{1}{c|}{} &
  \multicolumn{1}{c|}{MobileNet v2~\cite{mobilenet}} &
  \multicolumn{1}{c|}{} &
   &
  \multicolumn{1}{c|}{72.14} &
  \multicolumn{1}{c|}{71.61} &
  \multicolumn{1}{c|}{65.74} &
  \multicolumn{1}{c|}{53.50} &
  \multicolumn{1}{c|}{67.76} &
  \multicolumn{1}{c|}{53.46} &
  0.25 \\ \hline
\multicolumn{1}{|c|}{\begin{tabular}[c]{@{}c@{}}Speech \\ Recognition\end{tabular}} &
  \multicolumn{1}{c|}{Transformer} &
  \multicolumn{1}{c|}{Wav2Vec 2.0~\cite{wav2vec2}} &
  \multicolumn{1}{c|}{LibriSpeech} &
  WER ↓ &
  \multicolumn{1}{c|}{18.90} &
  \multicolumn{1}{c|}{18.83} &
  \multicolumn{1}{c|}{23.71} &
  \multicolumn{1}{c|}{21.99} &
  \multicolumn{1}{c|}{20.63} &
  \multicolumn{1}{c|}{21.98} &
  42.62 \\ \hline
\multicolumn{1}{|c|}{Recommendations} &
  \multicolumn{1}{c|}{MLPs} &
  \multicolumn{1}{c|}{DLRM~\cite{dlrm}} &
  \multicolumn{1}{c|}{\begin{tabular}[c]{@{}c@{}}Criteo\\ Terabyte\end{tabular}} &
  AUC ↑ &
  \multicolumn{1}{c|}{0.803} &
  \multicolumn{1}{c|}{0.803} &
  \multicolumn{1}{c|}{0.802} &
  \multicolumn{1}{c|}{0.801} &
  \multicolumn{1}{c|}{0.802} &
  \multicolumn{1}{c|}{0.801} &
  0.7947 \\ \hline
\end{tabular}%
}
\end{table}

%% file: Experiment_results/discriminative_error_diffusion.tex
\begin{table}[h]
\centering
\caption{Error diffusion for PTQ with MX data formats.
Both activations and pre-trained weights from the baseline model are quantized to the column's datatype.\\}
\label{tab:disc-ed}
\resizebox{0.78\textwidth}{!}{%
\renewcommand{\arraystretch}{1.4}
\begin{tabular}{ccccc|c|ccc|}
 \hline
\multicolumn{1}{|c|}{\multirow{2}{*}{Task}} &
  \multicolumn{1}{c|}{\multirow{2}{*}{Family}} &
  \multicolumn{1}{c|}{\multirow{2}{*}{Model}} &
  \multicolumn{1}{c|}{\multirow{2}{*}{Dataset}} &
  \multirow{2}{*}{Metric} &
  \multicolumn{1}{c|}{\multirow{2}{*}{\begin{tabular}[c]{@{}c@{}}FP32 \\ Baseline\end{tabular}}} &
  \multicolumn{2}{c|}{MXFP6} &
  \multirow{2}{*}{MXFP4} \\ \cline{7-8}
\multicolumn{1}{|c|}{} &
  \multicolumn{1}{c|}{} &
  \multicolumn{1}{c|}{} &
  \multicolumn{1}{c|}{} &
  \multicolumn{1}{c|}{} & &
  \multicolumn{1}{c|}{E2M3} &
  \multicolumn{1}{c|}{E3M2} &
   \\ \hline
\multicolumn{1}{|c|}{\multirow{5}{*}{\begin{tabular}[c]{@{}c@{}}Image \\ Classification\end{tabular}}} &
  \multicolumn{1}{c|}{\multirow{2}{*}{\begin{tabular}[c]{@{}c@{}}Vision \\ Transformer\end{tabular}}} &
  \multicolumn{1}{c|}{DeiT-Tiny~\cite{deit}} &
  \multicolumn{1}{c|}{\multirow{5}{*}{\begin{tabular}[c]{@{}c@{}}ImageNet\\ ILSVRC12\end{tabular}}} &
  \multirow{5}{*}{\begin{tabular}[c]{@{}c@{}}Top-1\\ Acc. ↑\end{tabular}} &
  \multicolumn{1}{c|}{72.16} &
  \multicolumn{1}{c|}{72.16} &
  \multicolumn{1}{c|}{71.29} &
  64.76 \\ \cline{3-3} \cline{6-9} 
\multicolumn{1}{|c|}{} &
  \multicolumn{1}{c|}{} &
  \multicolumn{1}{c|}{DeiT-Small~\cite{deit}} &
  \multicolumn{1}{c|}{} &
  &
  \multicolumn{1}{c|}{80.54} &
  \multicolumn{1}{c|}{80.50} &
  \multicolumn{1}{c|}{80.25} &
  76.80 \\ \cline{2-3} \cline{6-9} 
\multicolumn{1}{|c|}{} &
  \multicolumn{1}{c|}{\multirow{3}{*}{CNN}} &
  \multicolumn{1}{c|}{ResNet-18~\cite{resnet}} &
  \multicolumn{1}{c|}{} &
  &
  \multicolumn{1}{c|}{70.79} &
  \multicolumn{1}{c|}{70.66} &
  \multicolumn{1}{c|}{70.15} &
  67.40 \\ \cline{3-3} \cline{6-9} 
\multicolumn{1}{|c|}{} &
  \multicolumn{1}{c|}{} &
  \multicolumn{1}{c|}{ResNet-50~\cite{resnet}} &
  \multicolumn{1}{c|}{} &
  &
  \multicolumn{1}{c|}{77.40} &
  \multicolumn{1}{c|}{77.15} &
  \multicolumn{1}{c|}{76.48} &
  69.99 \\ \cline{3-3} \cline{6-9} 
\multicolumn{1}{|c|}{} &
  \multicolumn{1}{c|}{} &
  \multicolumn{1}{c|}{MobileNet v2~\cite{mobilenet}} &
  \multicolumn{1}{c|}{} &
  &
  \multicolumn{1}{c|}{72.14} &
  \multicolumn{1}{c|}{70.22} &
  \multicolumn{1}{c|}{65.32} &
  18.88 \\ \hline
\multicolumn{1}{|c|}{\begin{tabular}[c]{@{}c@{}}Speech \\ Recognition\end{tabular}} &
  \multicolumn{1}{c|}{Transformer} &
  \multicolumn{1}{c|}{Wav2Vec 2.0~\cite{wav2vec2}} &
  \multicolumn{1}{c|}{LibriSpeech} &
  WER ↓ &
  \multicolumn{1}{c|}{18.90} &
  \multicolumn{1}{c|}{19.09} &
  \multicolumn{1}{c|}{19.36} &
  24.39 \\ \hline
\end{tabular}
}
\end{table}

%% file: Experiment_results/discriminative_finetune.tex
\begin{table}[h]
\centering
\caption{Finetuned inference with MX data formats.
Finetuning is performed for a few epochs starting from the FP32 model.
Cells containing N/A means no finetuning was needed due to good direct-cast results.
MXFP6\_E2M3 achieves close-to-parity with FP32 after finetuning.\\}
\label{tab:disc-finetune}
\resizebox{0.8\textwidth}{!}{%
\renewcommand{\arraystretch}{1.4}
\begin{tabular}{ccccc|c|ccc|}
 \hline
\multicolumn{1}{|c|}{\multirow{2}{*}{Task}} &
  \multicolumn{1}{c|}{\multirow{2}{*}{Family}} &
  \multicolumn{1}{c|}{\multirow{2}{*}{Model}} &
  \multicolumn{1}{c|}{\multirow{2}{*}{Dataset}} &
  \multirow{2}{*}{Metric} &
  \multicolumn{1}{c|}{\multirow{2}{*}{\begin{tabular}[c]{@{}c@{}}FP32 \\ Baseline\end{tabular}}} &
  \multicolumn{2}{c|}{MXFP6} &
  \multirow{2}{*}{MXFP4} \\ \cline{7-8}
\multicolumn{1}{|c|}{} &
  \multicolumn{1}{c|}{} &
  \multicolumn{1}{c|}{} &
  \multicolumn{1}{c|}{} &
  \multicolumn{1}{c|}{} & &
  \multicolumn{1}{c|}{E2M3} &
  \multicolumn{1}{c|}{E3M2} &
   \\ \hline
\multicolumn{1}{|c|}{\multirow{3}{*}{\begin{tabular}[c]{@{}c@{}}Language \\ Translation\end{tabular}}} &
  \multicolumn{1}{c|}{\multirow{2}{*}{\begin{tabular}[c]{@{}c@{}}Transformers\\ (Enc-Dec)\end{tabular}}} &
  \multicolumn{1}{c|}{Transformer-Base~\cite{transformer}} &
  \multicolumn{1}{c|}{\multirow{2}{*}{WMT-17}} &
  \multirow{3}{*}{\begin{tabular}[c]{@{}c@{}}BLEU\\ Score ↑\end{tabular}} &
  \multicolumn{1}{c|}{26.85} &
  \multicolumn{1}{c|}{26.98} &
  \multicolumn{1}{c|}{27.01} &
  25.97 \\ \cline{3-3} \cline{6-9} 
\multicolumn{1}{|c|}{} &
  \multicolumn{1}{c|}{} &
  \multicolumn{1}{c|}{Transformer-Large~\cite{transformer}} &
  \multicolumn{1}{c|}{} &
  &
  \multicolumn{1}{c|}{27.63} &
  \multicolumn{1}{c|}{27.60} &
  \multicolumn{1}{c|}{27.62} &
  27.33 \\ \cline{2-4} \cline{6-9} 
\multicolumn{1}{|c|}{} &
  \multicolumn{1}{c|}{LSTM} &
  \multicolumn{1}{c|}{GNMT~\cite{gnmt}} &
  \multicolumn{1}{c|}{WMT-16} &
  &
  \multicolumn{1}{c|}{24.44} &
  \multicolumn{1}{c|}{N/A} &
  \multicolumn{1}{c|}{N/A} &
  24.56 \\ \hline
\multicolumn{1}{|c|}{\multirow{5}{*}{\begin{tabular}[c]{@{}c@{}}Image \\ Classification\end{tabular}}} &
  \multicolumn{1}{c|}{\multirow{2}{*}{\begin{tabular}[c]{@{}c@{}}Vision \\ Transformer\end{tabular}}} &
  \multicolumn{1}{c|}{DeiT-Tiny~\cite{deit}} &
  \multicolumn{1}{c|}{\multirow{5}{*}{\begin{tabular}[c]{@{}c@{}}ImageNet\\ ILSVRC12\end{tabular}}} &
  \multirow{5}{*}{\begin{tabular}[c]{@{}c@{}}Top-1\\ Acc. ↑\end{tabular}} &
  \multicolumn{1}{c|}{72.16} &
  \multicolumn{1}{c|}{72.09} &
  \multicolumn{1}{c|}{70.86} &
  66.41 \\ \cline{3-3} \cline{6-9} 
\multicolumn{1}{|c|}{} &
  \multicolumn{1}{c|}{} &
  \multicolumn{1}{c|}{DeiT-Small~\cite{deit}} &
  \multicolumn{1}{c|}{} &
  &
  \multicolumn{1}{c|}{80.54} &
  \multicolumn{1}{c|}{80.43} &
  \multicolumn{1}{c|}{79.76} &
  77.61 \\ \cline{2-3} \cline{6-9} 
\multicolumn{1}{|c|}{} &
  \multicolumn{1}{c|}{\multirow{3}{*}{CNN}} &
  \multicolumn{1}{c|}{ResNet-18~\cite{resnet}} &
  \multicolumn{1}{c|}{} &
  &
  \multicolumn{1}{c|}{70.79} &
  \multicolumn{1}{c|}{70.6} &
  \multicolumn{1}{c|}{69.85} &
  67.19 \\ \cline{3-3} \cline{6-9} 
\multicolumn{1}{|c|}{} &
  \multicolumn{1}{c|}{} &
  \multicolumn{1}{c|}{ResNet-50~\cite{resnet}} &
  \multicolumn{1}{c|}{} &
  &
  \multicolumn{1}{c|}{77.40} &
  \multicolumn{1}{c|}{77.27} &
  \multicolumn{1}{c|}{76.54} &
  74.86 \\ \cline{3-3} \cline{6-9} 
\multicolumn{1}{|c|}{} &
  \multicolumn{1}{c|}{} &
  \multicolumn{1}{c|}{MobileNet v2~\cite{mobilenet}} &
  \multicolumn{1}{c|}{} &
  &
  \multicolumn{1}{c|}{72.14} &
  \multicolumn{1}{c|}{71.49} &
  \multicolumn{1}{c|}{70.27} &
  65.41 \\ \hline
\multicolumn{1}{|c|}{\begin{tabular}[c]{@{}c@{}}Speech \\ Recognition\end{tabular}} &
  \multicolumn{1}{c|}{Transformer} &
  \multicolumn{1}{c|}{Wav2Vec 2.0~\cite{wav2vec2}} &
  \multicolumn{1}{c|}{LibriSpeech} &
  WER ↓ &
  \multicolumn{1}{c|}{18.90} &
  \multicolumn{1}{c|}{N/A} &
  \multicolumn{1}{c|}{21.46} &
  29.64 \\ \hline
\end{tabular}%
}
\end{table}

%% file: Experiment_results/generative_inference.tex
\begin{table}[h]
\centering
\caption{GPT3-175B direct-cast inference results. Higher is better for all tasks. Each number is given $\pm$ the bootstrap estimated standard deviation. We only experiment with the higher mantissa width variant of each format (i.e., MXFP8\_e4m3 and MXFP6\_e2m3) given that the results in Section 5.2 show these variants works better for direct-cast inference.\\}
\label{tab:gpt3-inference}
\resizebox{\textwidth}{!}{%
\renewcommand{\arraystretch}{1.4}
\begin{tabular}{c|cccccccc|}
  \hline
\multicolumn{1}{|c|}{Tasks} &
  \multicolumn{1}{c|}{FP32} &
  \multicolumn{1}{c|}{MXINT8} &
  \multicolumn{1}{c|}{MXFP8} &
  \multicolumn{1}{c|}{MXFP6} &
  \multicolumn{1}{c|}{\begin{tabular}[c]{@{}c@{}}MXFP6 Wt\\ MXFP8 Act\end{tabular}} &
  \multicolumn{1}{c|}{\begin{tabular}[c]{@{}c@{}}MXFP4 Wt\\ MXFP8 Act\end{tabular}} &
  \multicolumn{1}{c|}{\begin{tabular}[c]{@{}c@{}}MXFP4 Wt\\ MXFP6 Act\end{tabular}} &
  MXFP4 \\ \hline
\multicolumn{1}{|c|}{ARC easy ↑} &
  \multicolumn{1}{c|}{0.744 ± 0.009} &
  \multicolumn{1}{c|}{0.740 ± 0.009} &
  \multicolumn{1}{c|}{0.738 ± 0.009} &
  \multicolumn{1}{c|}{0.737 ± 0.009} &
  \multicolumn{1}{c|}{0.740 ± 0.009} &
  \multicolumn{1}{c|}{0.749 ± 0.009} &
  \multicolumn{1}{c|}{0.744 ± 0.009} &
  0.748 ± 0.010 \\ \hline
\multicolumn{1}{|c|}{ARC challenge ↑} &
  \multicolumn{1}{c|}{0.480 ± 0.015} &
  \multicolumn{1}{c|}{0.481 ± 0.015} &
  \multicolumn{1}{c|}{0.485 ± 0.015} &
  \multicolumn{1}{c|}{0.480 ± 0.015} &
  \multicolumn{1}{c|}{0.478 ± 0.015} &
  \multicolumn{1}{c|}{0.486 ± 0.015} &
  \multicolumn{1}{c|}{0.487 ± 0.015} &
  0.425 ± 0.014 \\ \hline
\multicolumn{1}{|c|}{Lambada ↑} &
  \multicolumn{1}{c|}{0.755 ± 0.006} &
  \multicolumn{1}{c|}{0.754 ± 0.006} &
  \multicolumn{1}{c|}{0.708 ± 0.006} &
  \multicolumn{1}{c|}{0.745 ± 0.006} &
  \multicolumn{1}{c|}{0.725 ± 0.006} &
  \multicolumn{1}{c|}{0.728 ± 0.007} &
  \multicolumn{1}{c|}{0.754 ± 0.006} &
  0.623 ± 0.007 \\ \hline
\multicolumn{1}{|c|}{College CS ↑} &
  \multicolumn{1}{c|}{0.360 ± 0.049} &
  \multicolumn{1}{c|}{0.340 ± 0.048} &
  \multicolumn{1}{c|}{0.350 ± 0.048} &
  \multicolumn{1}{c|}{0.350 ± 0.048} &
  \multicolumn{1}{c|}{0.340 ± 0.048} &
  \multicolumn{1}{c|}{0.340 ± 0.046} &
  \multicolumn{1}{c|}{0.320 ± 0.047} &
  0.240 ± 0.043 \\ \hline
\multicolumn{1}{|c|}{Int. law ↑} &
  \multicolumn{1}{c|}{0.504 ± 0.046} &
  \multicolumn{1}{c|}{0.537 ± 0.046} &
  \multicolumn{1}{c|}{0.455 ± 0.046} &
  \multicolumn{1}{c|}{0.521 ± 0.046} &
  \multicolumn{1}{c|}{0.463 ± 0.046} &
  \multicolumn{1}{c|}{0.331 ± 0.043} &
  \multicolumn{1}{c|}{0.347 ± 0.043} &
  0.298 ± 0.045 \\ \hline
\multicolumn{1}{|c|}{Jurisprudence ↑} &
  \multicolumn{1}{c|}{0.454 ± 0.049} &
  \multicolumn{1}{c|}{0.435 ± 0.048} &
  \multicolumn{1}{c|}{0.491 ± 0.048} &
  \multicolumn{1}{c|}{0.454 ± 0.048} &
  \multicolumn{1}{c|}{0.472 ± 0.049} &
  \multicolumn{1}{c|}{0.463 ± 0.048} &
  \multicolumn{1}{c|}{0.418 ± 0.048} &
  0.324 ± 0.045 \\ \hline
\end{tabular}%
}
\end{table}

\vspace{-1em}
\begin{table}[h]
\centering
\caption{LLaMA-7B direct-cast inference results. Higher is better for all tasks except \texttt{wikitext}. For this benchmark only, the Softmax function was not quantized to Bfloat16.\\}
\label{tab:llama-inference}
\resizebox{\textwidth}{!}{%
\renewcommand{\arraystretch}{1.4}
\begin{tabular}{c|cccccccc|}
\hline
\multicolumn{1}{|c|}{Tasks} &
  \multicolumn{1}{c|}{FP32} &
  \multicolumn{1}{c|}{MXINT8} &
  \multicolumn{1}{c|}{MXFP8} &
  \multicolumn{1}{c|}{MXFP6} &
  \multicolumn{1}{c|}{\begin{tabular}[c]{@{}c@{}}MXFP6 Wt\\ MXFP8 Act\end{tabular}} &
  \multicolumn{1}{c|}{\begin{tabular}[c]{@{}c@{}}MXFP4 Wt\\ MXFP8 Act\end{tabular}} &
  \multicolumn{1}{c|}{\begin{tabular}[c]{@{}c@{}}MXFP4 Wt\\ MXFP6 Act\end{tabular}} &
  MXFP4 \\ \hline
\multicolumn{1}{|c|}{ARC easy ↑} &
  \multicolumn{1}{c|}{0.729 ± 0.009} &
  \multicolumn{1}{c|}{0.725 ± 0.009} &
  \multicolumn{1}{c|}{0.716 ± 0.009} &
  \multicolumn{1}{c|}{0.718 ± 0.009} &
  \multicolumn{1}{c|}{0.726 ± 0.009} &
  \multicolumn{1}{c|}{0.697 ± 0.010} &
  \multicolumn{1}{c|}{0.696 ± 0.010} &
  0.637 ± 0.010 \\ \hline
\multicolumn{1}{|c|}{ARC challenge ↑} &
  \multicolumn{1}{c|}{0.447 ± 0.015} &
  \multicolumn{1}{c|}{0.444 ± 0.015} &
  \multicolumn{1}{c|}{0.430 ± 0.015} &
  \multicolumn{1}{c|}{0.445 ± 0.015} &
  \multicolumn{1}{c|}{0.442 ± 0.015} &
  \multicolumn{1}{c|}{0.412 ± 0.014} &
  \multicolumn{1}{c|}{0.406 ± 0.014} &
  0.355 ± 0.014 \\ \hline
\multicolumn{1}{|c|}{Lambada ↑} &
  \multicolumn{1}{c|}{0.736 ± 0.006} &
  \multicolumn{1}{c|}{0.731 ± 0.006} &
  \multicolumn{1}{c|}{0.720 ± 0.006} &
  \multicolumn{1}{c|}{0.724 ± 0.006} &
  \multicolumn{1}{c|}{0.721 ± 0.006} &
  \multicolumn{1}{c|}{0.675 ± 0.006} &
  \multicolumn{1}{c|}{0.678 ± 0.007} &
  0.557 ± 0.007 \\ \hline
\multicolumn{1}{|c|}{College CS ↑} &
  \multicolumn{1}{c|}{0.260 ± 0.044} &
  \multicolumn{1}{c|}{0.220 ± 0.045} &
  \multicolumn{1}{c|}{0.270 ± 0.042} &
  \multicolumn{1}{c|}{0.240 ± 0.043} &
  \multicolumn{1}{c|}{0.280 ± 0.045} &
  \multicolumn{1}{c|}{0.240 ± 0.043} &
  \multicolumn{1}{c|}{0.210 ± 0.041} &
  0.220 ± 0.042 \\ \hline
\multicolumn{1}{|c|}{Int. law ↑} &
  \multicolumn{1}{c|}{0.463 ± 0.046} &
  \multicolumn{1}{c|}{0.430 ± 0.045} &
  \multicolumn{1}{c|}{0.413 ± 0.045} &
  \multicolumn{1}{c|}{0.422 ± 0.045} &
  \multicolumn{1}{c|}{0.413 ± 0.045} &
  \multicolumn{1}{c|}{0.398 ± 0.045} &
  \multicolumn{1}{c|}{0.405 ± 0.045} &
  0.331 ± 0.041 \\ \hline
\multicolumn{1}{|c|}{Jurisprudence ↑} &
  \multicolumn{1}{c|}{0.361 ± 0.046} &
  \multicolumn{1}{c|}{0.370 ± 0.047} &
  \multicolumn{1}{c|}{0.380 ± 0.047} &
  \multicolumn{1}{c|}{0.370 ± 0.046} &
  \multicolumn{1}{c|}{0.352 ± 0.047} &
  \multicolumn{1}{c|}{0.269 ± 0.045} &
  \multicolumn{1}{c|}{0.296 ± 0.044} &
  0.269 ± 0.043 \\ \hline
\multicolumn{1}{|c|}{wikitext ↓} &
  \multicolumn{1}{c|}{9.488} &
  \multicolumn{1}{c|}{9.504} &
  \multicolumn{1}{c|}{9.768} &
  \multicolumn{1}{c|}{9.628} &
  \multicolumn{1}{c|}{9.683} &
  \multicolumn{1}{c|}{11.476} &
  \multicolumn{1}{c|}{11.147} &
  27.201 \\ \hline
\end{tabular}%
}
\end{table}

%% file: Experiment_results/gen_training.tex
\begin{table}[h]
\begin{minipage}[b]{0.3\linewidth}
    \centering
    \resizebox{\textwidth}{!}{%
    \renewcommand{\arraystretch}{1.2}
    \begin{tabular}{|l|c|c|c|}
    \cline{1-4}
    \multirow{2}{*}{Model} & \multirow{2}{*}{FP32} &   \multicolumn{2}{|c|}{MXFP6} \\
    \cline{3-4}
    \multicolumn{1}{|c|}{} & \multicolumn{1}{|c|}{} & E2M3 & E3M2\\ \hline
    GPT-20M  & 3.98 & 4.02 & 4.01 \\ \hline
    GPT-150M  & 3.30 & 3.33 & 3.32 \\ \hline
    GPT-300M  & 3.11 & 3.13 & 3.12 \\ \hline
    GPT-1.5B & 2.74 & 2.75 & 2.75 \\ \hline
    \end{tabular}%
    }
    \vspace{5pt}
    \caption{Language model loss for training from scratch using MXFP6\_E3M2 for weights, activations, and gradients.}
    \vspace{50pt}
    \label{tab:GPT-mxfp6-training}
\end{minipage}%
\hfill
\begin{minipage}[b]{0.7\linewidth}
  \centerline{\includegraphics[width=0.8\linewidth]{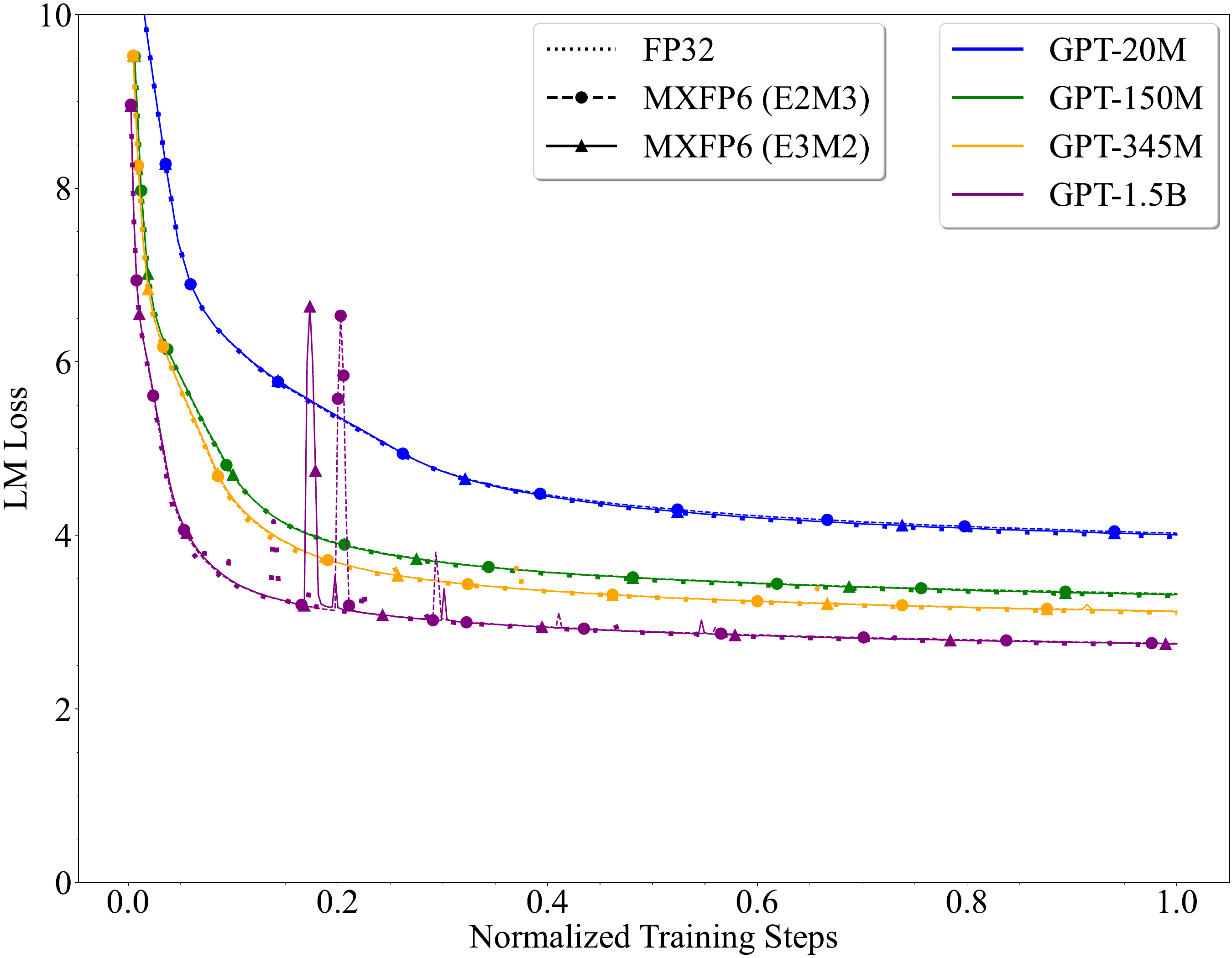}}
  \vspace{5px}
  \captionof{figure}{GPT training loss curve, using MXFP6\_E3M2 for weights, activations, and gradients.}
  \label{fig:gpt_mxfp6_training}
\end{minipage}%
\end{table}
\vspace{-2em}

%% file: Experiment_results/gen_training_mixed.tex
\begin{table}[ht]
\begin{minipage}[b]{0.3\linewidth}
    \centering
    \resizebox{\textwidth}{!}{%
    \renewcommand{\arraystretch}{1.2}
    \begin{tabular}{|l|c|c|}
    \cline{1-3}
    \multirow{2}{*}{Model} & \multirow{2}{*}{FP32} & MXFP4 Wt \\
    \multicolumn{1}{|c|}{} & \multicolumn{1}{|c|}{} & MXFP6 Act\\ \hline
    GPT-20M  & 3.98 & 4.04  \\ \hline
    GPT-150M  & 3.30 & 3.33  \\ \hline
    GPT-300M  & 3.11 & 3.14  \\ \hline
    GPT-1.5B & 2.74 & 2.76 \\ \hline
    \end{tabular}%
    }
    \vspace{5pt}
    \caption{Language model loss for training from scratch using MXFP4 for weights and MXFP6\_E3M2 for activations and gradients.}
    \vspace{50pt}
    \label{tab:GPT-mixed_training}
\end{minipage}%
\hfill
\begin{minipage}[b]{0.7\linewidth}
  \centerline{\includegraphics[width=0.8\linewidth]{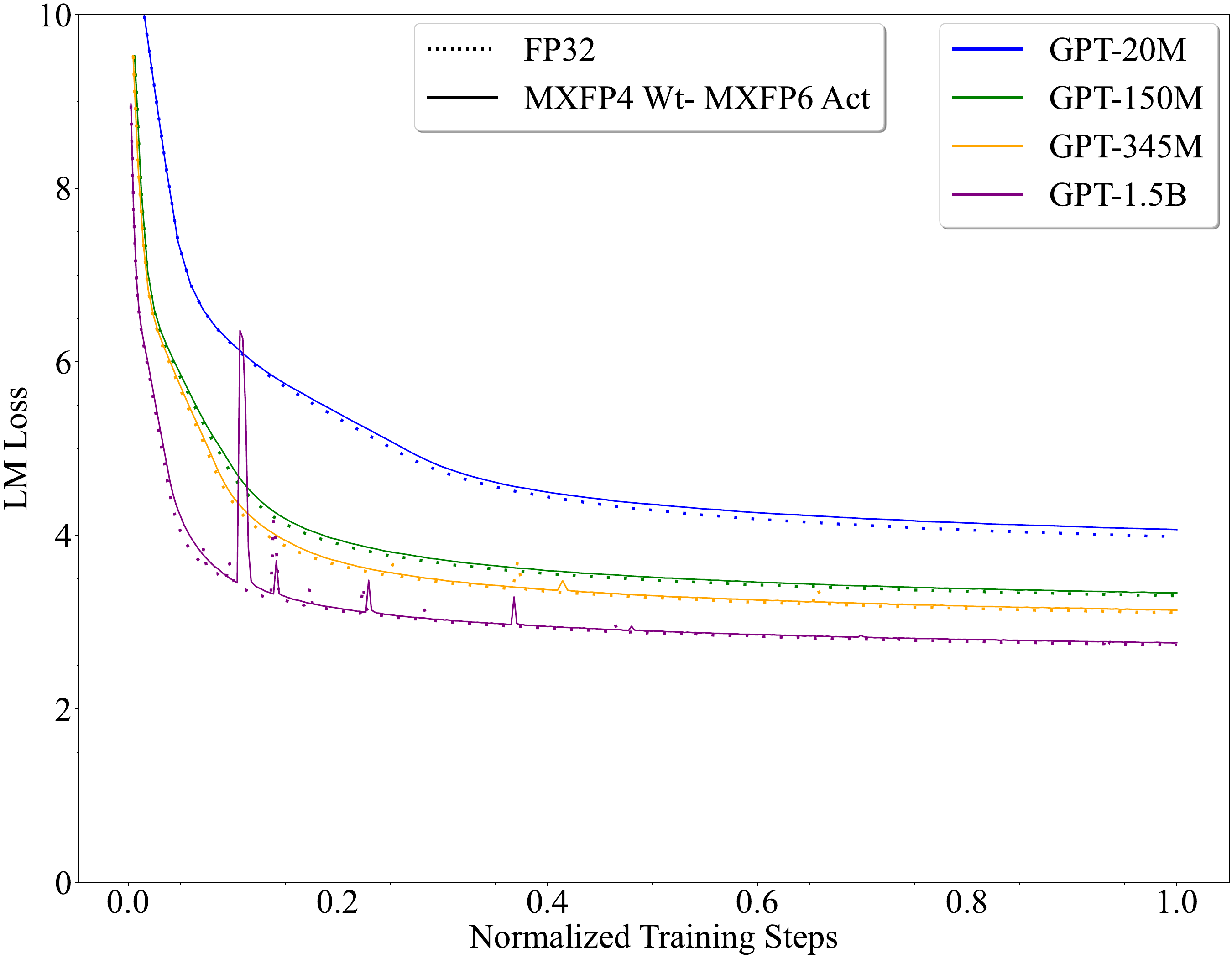}}
  \vspace{5px}
  \captionof{figure}{GPT mixed-precision training loss curve, using MXFP4 for weights and MXFP6\_E3M2 for activations and gradients.}
  \label{fig:gpt_mixed_training}
\end{minipage}%
\end{table}